\documentclass[conference]{IEEEtran}
\IEEEoverridecommandlockouts
\usepackage{cite}
\usepackage{amsmath,amssymb,amsfonts}
\usepackage{algorithmic}
\usepackage{graphicx}
\usepackage{textcomp}
\usepackage{xcolor}
\usepackage{amsmath,amsfonts}
\usepackage{bbm}
\usepackage{url}
\usepackage{hyperref}
\usepackage{amssymb}
\usepackage{rotating}
\usepackage{booktabs}
\usepackage{pifont}
\usepackage{multirow}
\usepackage[flushleft]{threeparttable}
\usepackage{color}
\usepackage{listings}
\usepackage{array}
\usepackage{arydshln}
\usepackage{graphicx}
\usepackage{enumitem}
\usepackage{pdflscape}
\usepackage[dvipsnames]{xcolor}
\usepackage{graphicx}
\usepackage{amsmath}
\usepackage{titlesec}
\usepackage{tikz}
\usetikzlibrary{arrows.meta}
\usepackage{subcaption}
\def\BibTeX{{\rm B\kern-.05em{\sc i\kern-.025em b}\kern-.08em
    T\kern-.1667em\lower.7ex\hbox{E}\kern-.125emX}}

\begin{document}

\title{
\vspace{0.8cm}
\LARGE \bf 
Multi-Pedestrian Safety Warning at Urban Intersections:\\
Use Case of Digital Twin 
%Safety Assessment

}

\author{Yongjie Fu, Qi Gao, 
Mahshid Ghasemi Dehkordi, Gil Zussman, Xuan Di$^*$
\thanks{This work has been accepted by IEEE ITSC 2026.}
\thanks{This work was supported by NSF CPS-2038984, ERC-2133516.}
\thanks{$*$Corresponding author: Xuan Di (Email: sharon.di@columbia.edu).}
\thanks{Yongjie Fu, Qi Gao, and Xuan Di are with the Department of Civil Engineering and Engineering Mechanics at Columbia University, New York. Xuan Di is also with the Data Science Institute. Email:\{yf2578, qg2179, sharon.di\}@columbia.edu}
\thanks{Mahshid Ghasemi Dehkordi and Gil Zussman are with the Department of Electrical Engineering at Columbia University, New York. Email: \{mg4089, 	gil.zussman\}@columbia.edu}
}

\maketitle


\begin{abstract}
 Digital twins (DTs) for urban transportation systems have gained increasing attention; however, their systematic evaluation in safety-critical scenarios remains limited. This paper presents a multi-pedestrian safety warning system at urban intersections enabled by a tightly coupled physical–digital twin framework. Built upon the COSMOS city-scale wireless testbed in New York City, the proposed system integrates camera and ultra-wideband (UWB), edge-cloud computing, predictive trajectory modeling, and MQTT-based communication to deliver real-time safety alerts to vulnerable road users (VRUs). The system is evaluated through both field deployment and virtual reality (VR) experiments. Results demonstrate high warning generation accuracy, localization accuracy, efficient end-to-end latency under different model configurations, and significant reductions in user response time when warnings are issued. The proposed DT framework provides a scalable, modular, and generalizable solution for real-time multi-pedestrian safety enhancement at complex urban intersections.

\end{abstract}


% \fi
\IEEEpeerreviewmaketitle

% % \newcommand{\gz}[1]{\textcolor{red}{[Gil: #1]}}
% % \newcommand{\zk}[1]{\textcolor{lime}{[Zoran: #1]}}
% % \newcommand{\mt}[1]{\textcolor{blue}{[Mehmet: #1]}}
% % \newcommand{\md}[1]{\textcolor{cyan}{[Mahshid: #1]}}
% % \newcommand{\sd}[1]{\textcolor{magenta}{[Sharon: #1]}}
% % \newcommand{\yf}[1]{\textcolor{grey}{[Yongjie: #1]}}
% }

\vspace{-.3cm}
\section{Introduction}
\label{sec-1-r3}
\vspace{-.1cm}

There is a growing body of literature on digital twin (DT), while its evaluation is understudied, especially for safety-critical scenarios such as pedestrian-vehicle interaction at urban intersections.
This paper aims to develop a multi-pedestrian safety warning system at urban intersections, relying on video analytics, wireless communication, Ultra-wideband (UWB) localization, and MQTT messaging. 
Its efficiency is evaluated in both field experiments and VR environments.

Vulnerable road users (VRUs) refer to non-motorists such as pedestrian, bicyclists, other cyclists, or persons on personal conveyance~\cite{fhwa22vru}.

The studies of DTs for urban traffic management, especially involving VRUs, are lacking, 
%The literature on DTs for urban traffic management is lacking, 
partly because the development of a DT for a system is non-trivial, particularly when involved with humans. 
% Accordingly, we strongly believe that a comprehensive review is warranted on trans-disciplinary topics. %at the intersection of transportation, DT, and CPS.
%%%%%%%%%%%%%%%%%%%%%%%%%%%%%%%%%%%%%%%%%%%%%%%%%%%%%
%This paper presents challenges and opportunities of interdisciplinary collaboration for the development of urban transportation DT system from the perspectives of methods and applications. %, particularly in safety-time critical applications like pedestrian vehicle collisions. 
%This paper presents a survey paper on methods and applications of DT for urban traffic management. 
This paper presents a multi-person safety warning system based on a DT architecture. We utilize a DT for real-time traffic monitoring, prediction, and warning using an existing physical testbed deployed in New York City (NYC), leveraging advanced sensing, communication, computing, and AI-based automation technologies. We evaluate the system in terms of UWB-based localization accuracy, end-to-end system latency, and VR user experience.

% The overall contributions of this paper include: 
% (1) present the application design; 
% (2) propose a prototype for reference; 
% and (3) validate the system in terms of localization accuracy, warning accuracy, latency, and user response.

The rest of this paper is organized below. Section.~\ref{sec:related-work} reviews the related work. 
%sensing and communication technologies, 
% object detection and tracking, 
% and urban transportation use cases. 
Section.~\ref{sec:ourDT} demonstrates the architecture of our DT, building on a real-world testbed. Section~\ref{sec:method} introduces the technical details of the safety warning implementation. Section.~\ref{subsec:our_eval} presents the evaluation results in both virtual reality (VR) and real-world settings.
Section.~\ref{sec:conclud} concludes our work and presents future work.

\begin{figure*}[h!]
\centering

\begin{subfigure}{0.64\textwidth}
    \centering
    \includegraphics[width=\textwidth]{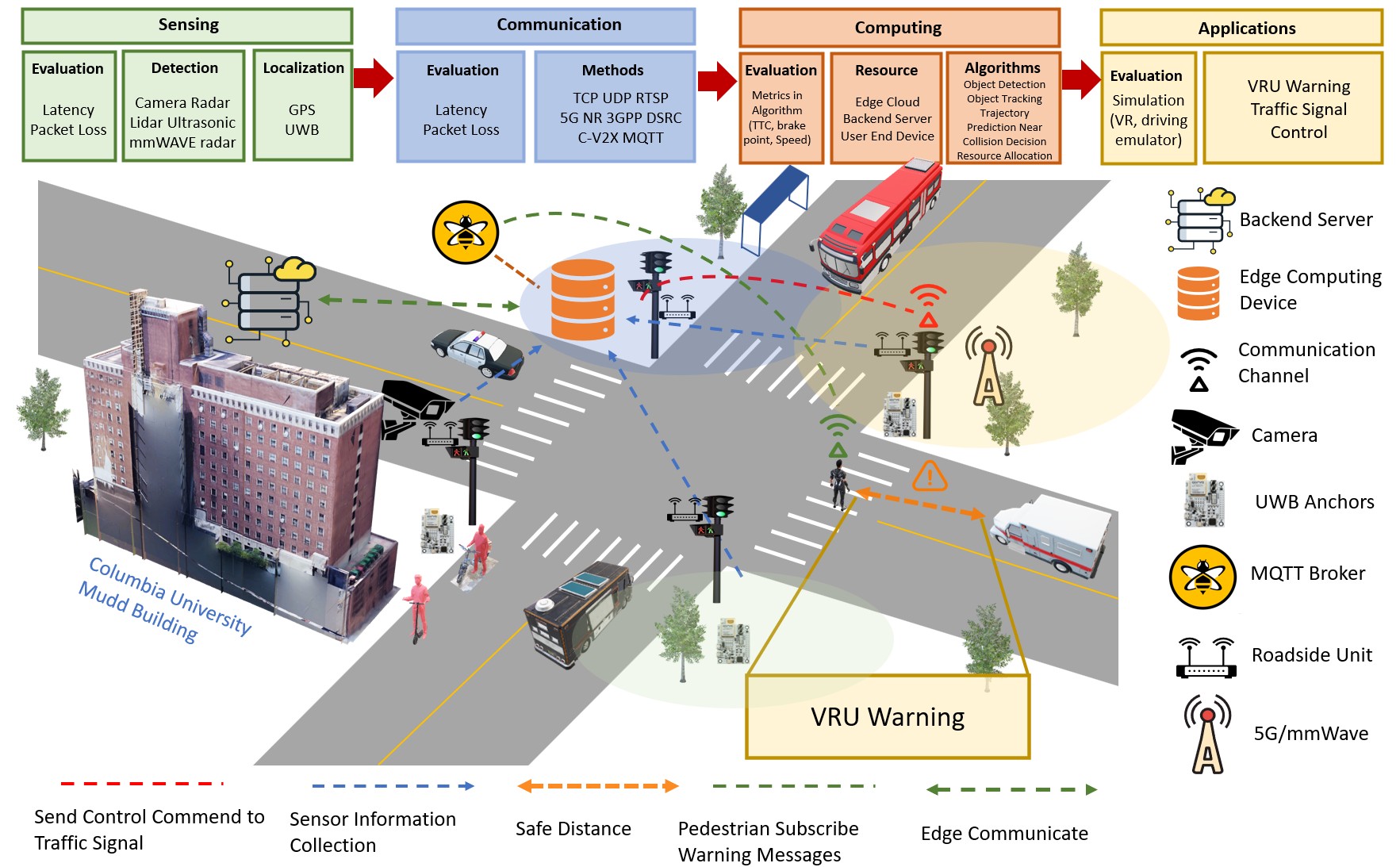}
    \caption{Architecture of the proposed DT pipeline.}
    \label{fig:dt_overview}
\end{subfigure}
\hfill
\begin{subfigure}{0.34\textwidth}
    \centering
    \includegraphics[width=\textwidth]{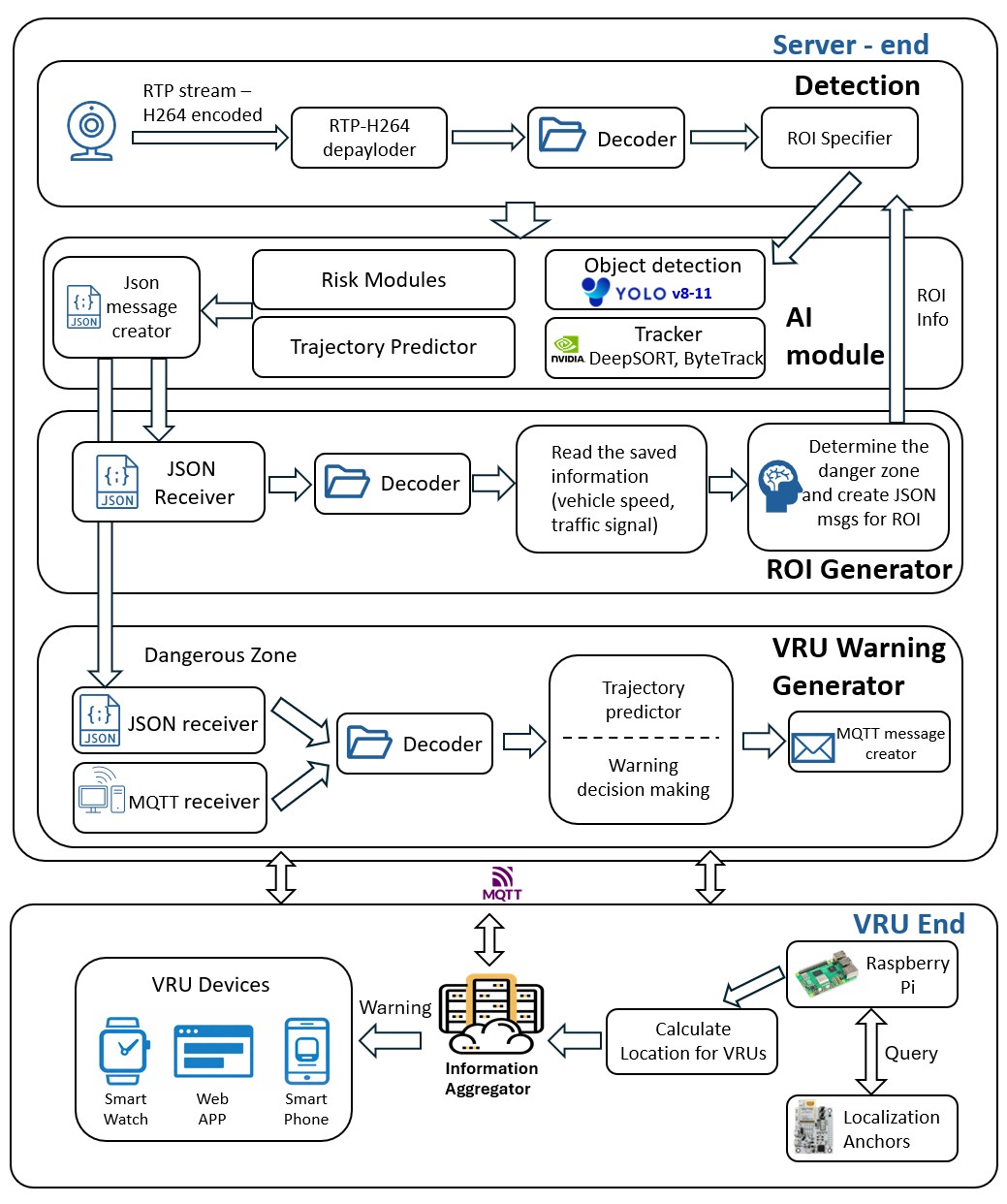}
    \caption{Diagram of intersection safety warning.}
    \label{fig:safety-ROI-app}
\end{subfigure}

\caption{Overview of the proposed digital twin pipeline and its intersection safety warning application.}
\label{fig:dt_pipeline_safety_app}
\vspace{-0.4cm}
\end{figure*}

\section{Related work}
\label{sec:related-work}
\subsection{Intersection Pedestrian Safety Warning}
Infrastructure-based pedestrian safety warning systems at urban intersections typically rely on camera, LiDAR, or radar sensing to detect and track VRUs. Vision-based approaches leverage deep learning models for object detection and tracking, followed by time-to-collision (TTC) or distance-based risk assessment to trigger warnings \cite{Ghasemi2025VideoAnalytics, salzmann2020trajectron}. Such systems have demonstrated effectiveness in real-time traffic monitoring; however, they are often reactive and sensitive to occlusion, lighting variations, and sensing noise. Other studies integrate multi-sensor perception to improve robustness in dense urban environments, yet most focus primarily on detection accuracy rather than predictive risk modeling or closed-loop safety intervention \cite{wang2024smartmobility}. Furthermore, these systems generally operate as standalone perception modules without a synchronized digital representation that enables trajectory simulation, threshold optimization, or human behavioral evaluation.

\subsection{Device-Based Pedestrian Localization and Warning}
UWB localization enables centimeter-level pedestrian tracking and has been increasingly adopted for safety monitoring due to its robustness and low power consumption. Prior work addresses practical challenges such as human body shadowing and NLOS effects through fusion and adaptive filtering techniques~\cite{Han2020Cooperative}. UWB has also been applied to proximity-based warning systems in industrial and construction environments~\cite{Yu2021Substation, Mastrolembo2023PWS}. 

While these studies demonstrate the feasibility of high-precision localization for safety alerts, they primarily focus on standalone positioning or proximity warnings in controlled settings. Integration with multi-modal infrastructure sensing, predictive trajectory modeling, and synchronized digital twin frameworks for complex urban intersections remains limited.

\subsection{Digital Twin–Enabled Urban Safety Systems}

Recent work has demonstrated the feasibility of deploying real-time pedestrian safety systems using edge video analytics at urban intersections. Fu \textit{et al.}~\cite{fu2024digital,fu2026ai} proposed a digital twin-based warning framework that integrates object detection, transformer-based trajectory prediction, and TTC computation within the COSMOS testbed. Similarly, Ghasemi \textit{et al.}~\cite{Ghasemi2025VideoAnalytics} developed PAVE, a scalable edge–end architecture that processes multi-camera video streams to identify danger zones and deliver privacy-preserving alerts to pedestrians.

While these systems achieve low-latency warning through infrastructure-based sensing and edge computing, they primarily rely on vision-based localization and predefined danger zones. Fine-grained multi-user positioning and synchronized physical–digital state alignment remain limited. In contrast, our work integrates centimeter-level multi-user UWB localization with camera sensing in the physical layer and embeds predictive trajectory modeling and safety computation in the digital twin layer, enabling a tightly coupled and bidirectional physical–digital safety framework. Main contributions of this work:
\begin{itemize}
    \item Develop a DT-enabled VRU safety warning system that integrates real-time sensing, communication, and analytics to support safety-critical intersection applications.
    
    \item Propose a multi-user UWB-based localization framework that enables accurate and concurrent tracking of multiple VRUs in dense urban environments.
    
    \item Validate the proposed system through both real-world and simulation-based experiments, evaluating (i) end-to-end system latency, (ii) localization and warning accuracy, and (iii) user-end behavioral responses to safety alerts.
\end{itemize}

\section{DT architecture}\label{sec:ourDT}

% \tcr{\textbf{
% Mahshid: POI;
% Mehmet: Obj. detection traj. prediction;
% Yongjie: safety warning. How it's tied back to resource allocation and upstream technological enablers?
% All: how these pieces are tied toward one pipeline}}
% \subsection{Dashboard}
% \tcr{\textbf{Kentyou: dashboard image, mechanism/flowchart, snapshots}}

% \begin{figure}[h]
% \centering
% \includegraphics[width=2.2 \columnwidth]{figs/DT_overview.pdf}
% \caption{Urban traffic digital twin.} \vspace{-1.2em}
% \label{fig:dt_overview}
% \end{figure}
%% !!

% In this section, we present a cyberphysical system enabled, vision based, publisher-subscriber architecture DT, 
% named ``\textbf{UrbanSafeDT}," 
% %\textbf{INTERStellar} (Safe Smart City Intersections as Intelligence Nodes for Future Metropolises). 
% %Interstellar 
% which senses VRUs at urban intersections, makes predictions of their movements, applies traffic operation and control strategies, and provides feedback to system controllers and road users (Fig.~\ref{fig:system_overall}), with the primary goal of increasing the safety of VRUs. 

In this section, we present a DT architecture for UT-DT, based on the sensing/communication/computing testbed deployed in NYC  (Fig.~\ref{fig:dt_overview}). The proposed architecture is enabled by 
%user software defined radios (SDR), 
cameras and LiDARs, 
high speed communications, 
and edge cloud computing. 

% \begin{figure}[h!]
% \centering
% \includegraphics[width=.9 \columnwidth]{figs/V_Ours/safety-ROI-diagram-v2.jpg}
% \caption{Diagram of intersection safety warning use case.}
% \label{fig:safety-ROI-app}
% \vspace{-0.4cm}
% \end{figure}

%\sd{Our DT is designed to be modular, in which each component is integrated into another via information flow, and to some degree we can decouple each module for testing and validation.}

% \vspace{-.3cm}
% \subsection{System components}
% \vspace{-.1cm}
% Bidirectional interaction between the physical and the digital
% The physical impacts the digital, while the digital reshapes the physical. 
% Below we will introduce each component and how they interact with each other.

\vspace{-.3cm}
\subsection{Physical Infrastructure}
\label{subsec:phys}
\vspace{-.1cm}

\begin{figure}[h!]
\centering
\includegraphics[width=.95 \columnwidth]{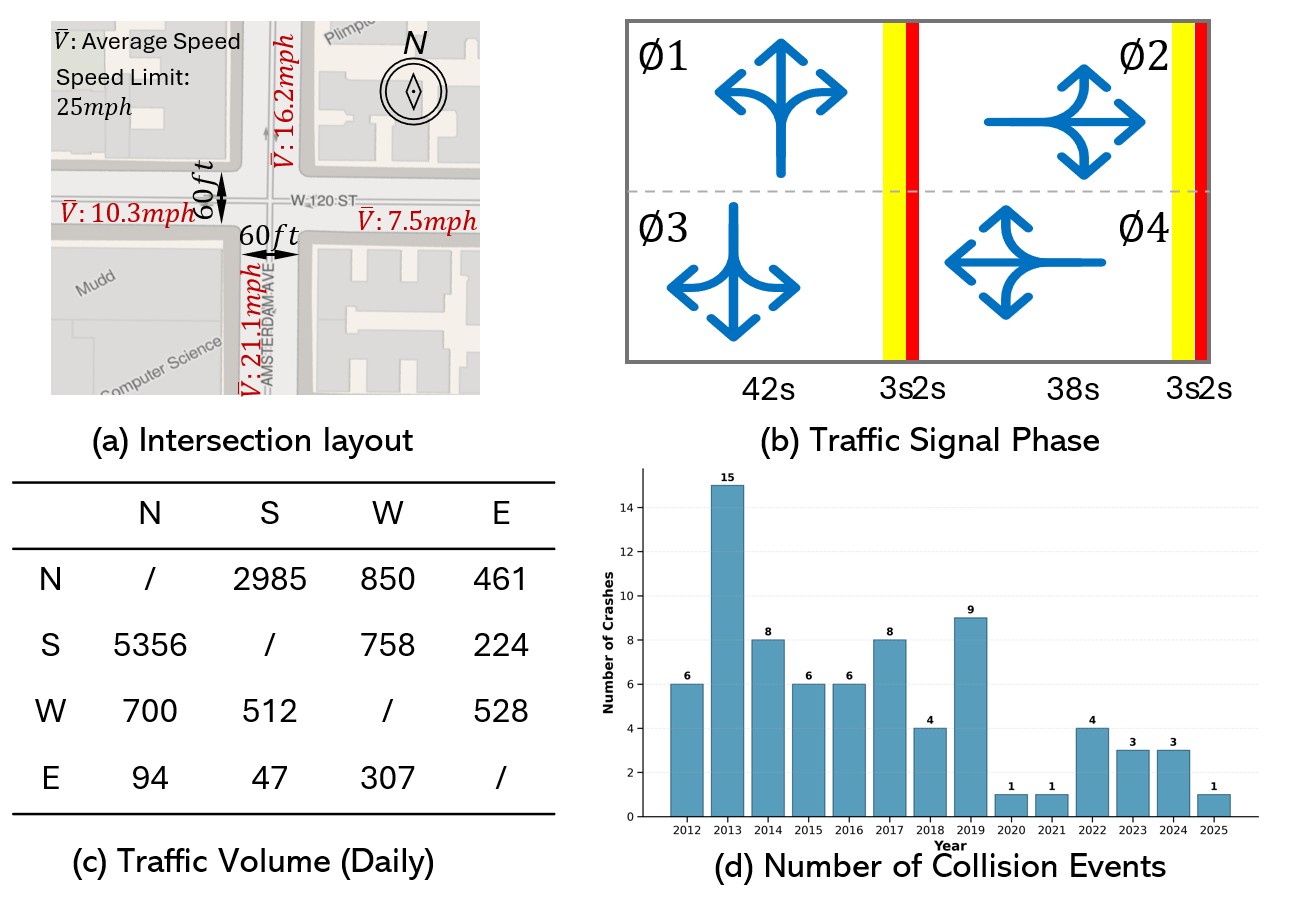}
\caption{Traffic statistics for NYC's intersection of 120th St. \& Amsterdam Ave.}
\label{fig:intersection-120}
\vspace{-0.4cm}
\end{figure}

Pilot experiments are executed at the signalized intersection of Amsterdam Avenue and 120th Street near Columbia campus in NYC. 
The road geometry and traffic statistics are summarized in Fig.~\ref{fig:intersection-120}.
% \rev{Both streets are two-lane bidirectional, with a lane-width of ...
% Along Amsterdam Ave, there is a left turning bay of length ?? m with an additional left-turning lane. 
% The traffic signal control is fixed-timing, with a cycle length of ?? seconds (consisting of ??-second green, ??-second yellow, ??-second red, and ??-second all red. 
% The daily traffic volumes are .... veh/hour/lane, 
% \sd{(summarize in a table?)}}

% \subsubsection{Road and traffic characteristics}
% \label{subsec:cosmos}
%\sd{Yongjie/Mehmet/Mahshid: could we include a map of this intersection, traffic statistics for the intersection: traffic volumes (car, ped.,micromobility), ave. spd, \# collisions (monthly or quaterly), ...}

\subsubsection{Technological enabler}
\label{subsec:cosmos}

% Our physical world is built on the COSMOS testbed (``Cloud enhanced Open Software defined mobile wireless testbed for city-Scale deployment”), %, funded by National Science Foundation Platforms for Advanced Wireless Research, 
% developed for the design, development, and deployment of an advanced wireless city-scale in order to support real-world experimentation on next-generation wireless technologies~\cite{raychaudhuri2020challenge}. 
% It targets the technology ``sweet spot” of ultra-high bandwidth and ultra-low latency, a capability that will enable a broad class of safety-time critical applications. 

The physical functionality of the proposed UT-DT is built on top of the COSMOS testbed (``Cloud enhanced Open Software defined mobile wireless testbed for city-Scale deployment”)~\cite{raychaudhuri2020challenge},
developed for real-world research, development, and deployment of city-scale advanced wireless communications and innovative applications. It targets the technology ``sweet spot” of ultra-high bandwidth and ultra-low latency, a capability that will enable a broad class of safety-time critical applications. 
%Realization of these applications involves not only faster radio links, but also aspects such as spectrum use, networking, and edge computing. 
%Edge cloud technology is integrated to support low-latency applications. 
%The testbed has been equipped with sensors, including cameras and lidars at intersections.
Deployed in West Harlem, NYC, next to Columbia University campus, the COSMOS testbed is 
%a convenient test site for us to 
an enabler for research on the
design, development, and deployment of a DT for urban traffic management. 

At the intersection of Amsterdam Avenue and 120th Street, cameras, LiDAR, 
and wireless sensing and communication nodes are deployed. 
The cameras employ H264 encoding with an I-frame interval set to 10 frames. The live RTSP video streams with 4K resolution at 30\thinspace{fps} were processed on a COSMOS edge server equipped with an A100 GPU.

\vspace{-.3cm}
\subsection{Digital twin}
\vspace{-.1cm}

% object detection and tracking

% prediction

% simulation

Building on data collected from the physical world, 
%we create a digital world using computer simulation.
%\sd{Mehmet: Unreal Engine for Simulator, cross-ref AxiV paper, insert a pic} \mt{Addressed in section "Synthetic Data Generation"}
we built an Unreal Engine based simulation platform %, called Boundless, 
for generating photorealistic street scenes. %, adapting the City Sample project released by Epic Games \cite{turkcan2024boundless}. 
% The improvements in rendering and raytracing allow for the realism gap in existing methods to be closed, which rely on older graphical assets (such as CARLA) or mobile-optimized game engines like Unity. By creating a DT of a COSMOS testbed site, we explored the suitability of this DT for object detection, achieving results competitive with real-world datasets. 
% We expect that with the increasing efforts on building digital twins of entire cities, this type of photorealistic simulation will soon enable training of more accurate object detection and tracking models, as well as their deployment.
Moving objects are populated into an integrated SUMO-CARLA simulation platform to validate our safety warning application \cite{fu2024digital},
%The digital world is useful for conducting experiments to validate the system, especially when some field tests are difficult to carry out in the real world. 
where vehicle locations are synchronized from the real-world MQTT messages. 

\subsection{VRU safety warning}\label{subsec:use}
\vspace{-.1cm}

Intersections, where sixty percent of crashes happen \cite{2024_fhwa}, are critical bottlenecks of an urban transportation network. 
To improve urban road safety and increase traffic capacities, safety warning is the key.
%safety warning and traffic signal optimization are the keys. 
Leveraging existing sensors at the intersection, we have developed a combined VRU and vehicle warning application.  
The cameras sense the presence of VRUs at urban intersections,
make predictions of their movements, apply traffic operation and control strategies, and feedback to system
controllers and road users, with the primary goal of
increasing traffic safety and efficiency 
%that issues warning messages to both VRUs and drivers simultaneously when a near crash is detected 
(see Fig.~\ref{fig:safety-ROI-app}). %The warning messages sent to pedestrians follow the MQTT protocol or via  5G/wireless, and those to drivers could be sent via fast dedicated channels such as 5G/wireless or C-V2X communication channels.
%The prototype of the safety warning app is tested at the intersection near the Columbia campus, leveraging the COSMOS camera, edge cloud computing, and communication. \\

%\subsubsection{Resource optimization}
\section{Methodology}
\label{sec:method}

\subsection{Resource optimization}

Due to the computational needs of complex computer vision models and the high volume of video data, ensuring scalability in a video analytics pipeline requires optimization of its configuration to maintain performance. However, the system's performance is impacted by factors such as the video content, network conditions, and available computational resources. As a result, it is crucial to implement a dynamic, real-time optimization mechanism that adjusts key configuration parameters—such as resolution, frame rate, and bitrate—based on these varying conditions. This adaptive approach allows the system to continuously balance performance with resource efficiency, ensuring scalable and reliable video analytics. Accordingly, we have equipped our video analytics pipeline with a \emph\textbf{Resource Optimization} (see Fig.~\ref{fig:dt_overview}) component that continuously adapts the system's configuration parameters to maintain its performance. 

In addition to dynamic adaptation of the system's configuration, we can reduce latency and GPU consumption by identifying Regions of Interest (RoIs) where pedestrians might be in danger, using lightweight processing (e.g., low resolution, small models). As shown in Table~\ref{tab:pipeline-elem-latency}, smaller models (``YOLOv8s" in Column 4) incur significantly lower latency. 
Larger models (``YOLOv8x" in Column 6) with higher resolution and frame rate are only triggered when a critical danger area, i.e., RoI, is detected. For example, detecting large objects such as vehicles does not require large models or high resolution. Therefore, we can use a smaller model and lower resolution to detect vehicles and their trajectories, and based on that, determine the danger areas where pedestrians may be at risk or in the blind zones of vehicles. These identified danger areas are then processed using larger models and higher resolution to detect pedestrians at risk and notify them or the vehicles if necessary. In our intersection safety warning system (shown in Fig.~\ref{fig:safety-ROI-app}), we have embedded an RoI Specifier element to facilitate this approach.

\subsection{Trajectory prediction model}

\begin{figure}[h!]
\centering
% --- Figure ---
\includegraphics[width=0.55\columnwidth]{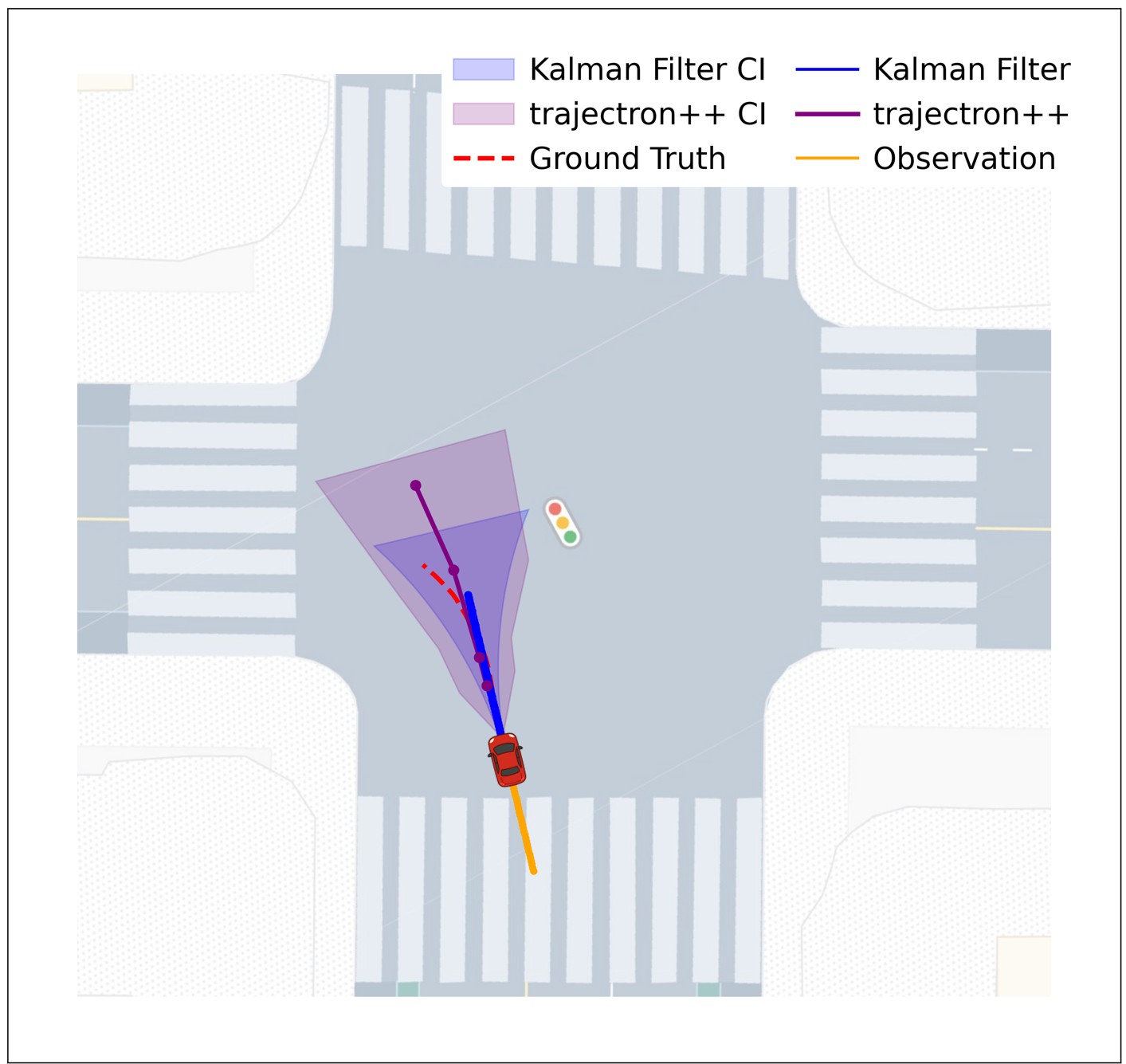}

\vspace{0.5em} % space between figure and table

% --- Table ---
\begin{tabular}{lcc}
    \hline
    Method & ADE $\downarrow$ & FDE $\downarrow$ \\
    \hline
    Kalman Filter & 0.91 & 1.92 \\
    Trajectron++  & \textbf{0.77} & \textbf{1.34} \\
    \hline
\end{tabular}

% --- Table title below table ---
\vspace{0.3em}
{ADE and FDE comparison. Lower values are better ($\downarrow$).}
\caption{Performance comparison for trajectory prediction}
\label{fig:traj-UQ}
\end{figure}

A variety of AI models are developed to predict future trajectories of interacting road users at the intersection, including 
InfoSTGCAN \cite{ruan2024infostgcan}, PI-NeuGODE, \cite{mo2024pi}, 
and uncertainty quantification (UQ) to characterize the predictive confidence \cite{mo2022uncertainty}. The Kalman Filter (KF) \cite{Ghasemi2025VideoAnalytics} performs a prediction by recursively estimating the state of a moving object using a linear dynamic model with Gaussian noise. Trajectron++ \cite{salzmann2020trajectron++} models multimodal future trajectories by combining recurrent neural networks with conditional variational autoencoders, enabling probabilistic and socially-aware predictions. We present the predicted trajectories and performance comparison in Fig.~\ref{fig:traj-UQ}, where the performance is evaluated using Average Displacement Error (ADE, i.e., the mean Euclidean distance between predicted and ground-truth positions over all future time steps) and Final Displacement Error (FDE, i.e., the Euclidean distance at the final prediction step).

\subsection{UWB Localization Method}

\begin{figure}[h!]
\centering
\includegraphics[width=.9 \columnwidth]{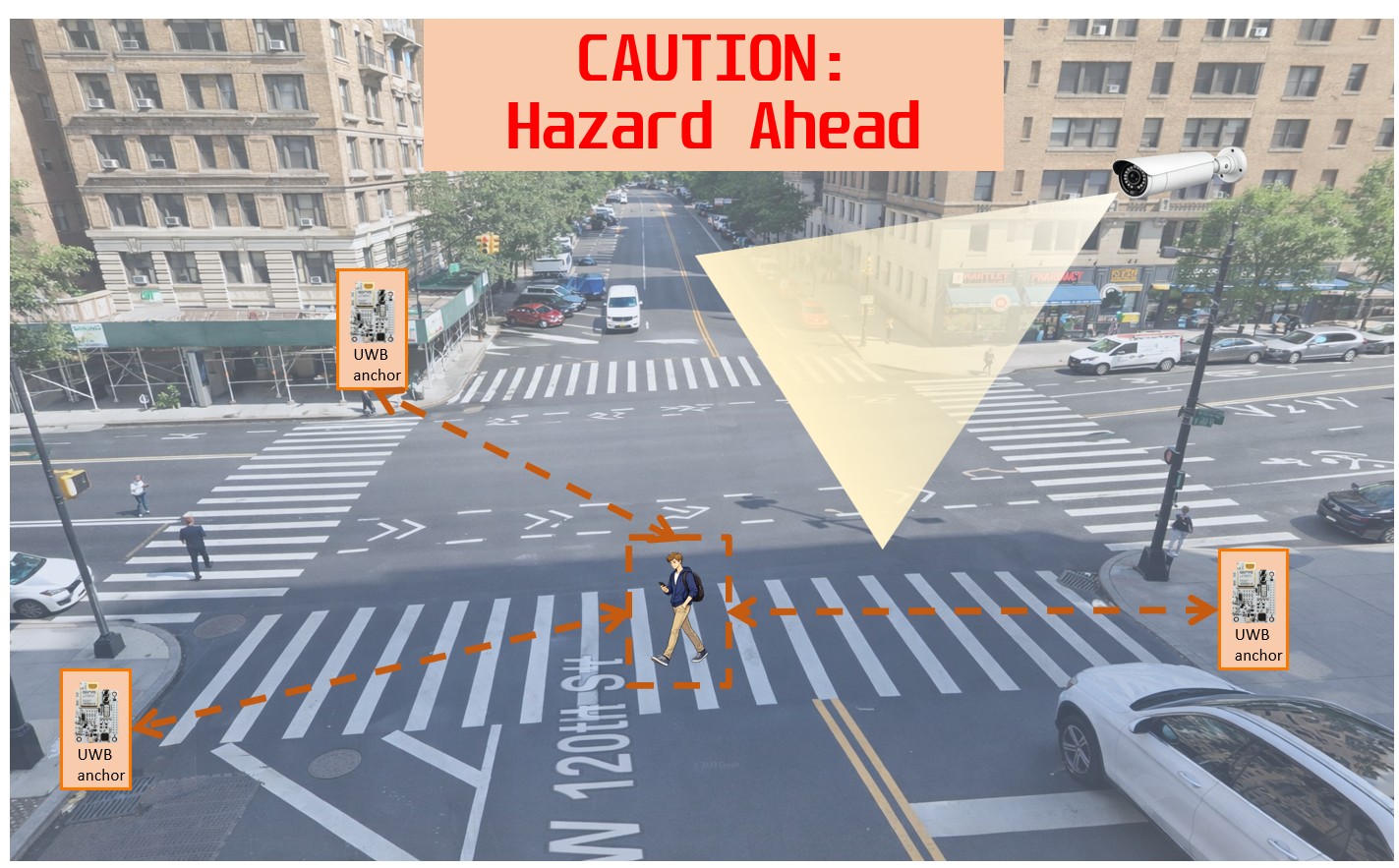}
\caption{Demonstration of UWB localization}
\label{fig:uwb}
\vspace{-0.2cm}
\end{figure}

We employ UWB technology for pedestrian localization using the native Two-Way Ranging (TWR) protocol. TWR estimates distance from the round-trip signal time between anchors and users without requiring clock synchronization. Under line-of-sight conditions, it achieves sub-10 cm accuracy in a plug-and-play configuration, making it suitable for on edge localization.

Three fixed UWB anchors are deployed along the roadside, operating in responder mode of CLI firmware (As shown in Fig.~\ref{fig:uwb}). Each pedestrian carries a UWB tag driven by Raspberry Pi, which acts as the initiator using Qorvo’s UCI firmware and toolkit to query all anchors. The 2D pedestrian position is then estimated via least-squares multilateration from the measured ranges.

Since native TWR only supports a single-initiator–multiple-anchor setup, we introduce a time-division scheduling mechanism for multi-user localization. A cyclic schedule of length \(NT\) is constructed, where \(T\) is the minimum window required for reliable ranging. Each user is assigned a dedicated slot of duration \(T\) to perform multiple ranging exchanges, improving robustness and accuracy. Outside its slot, the user remains silent to prevent channel contention, ensuring scalable and collision-free operation.

Assuming approximately constant pedestrian velocity over short intervals, we apply a constant-velocity model to interpolate positions outside the ranging window. Velocity is estimated from the two most recent measurements and used to propagate the state forward, providing continuous trajectory estimates between updates.

%\zm{To account for the heterogeneity of road users, we propose InfoSTGCAN \cite{ruan2024infostgcan}, which embeds road user trajectories into quantized latent codes. To further ground our trajectory prediction model in physical principles, we adopt the physics-informed deep learning (PIDL) \cite{mo2024pi,mo2021physics} technique. We also combine uncertainty quantification (UQ) with PIDL \cite{mo2022uncertainty} for worst-case warnings.}

%Fig.~\ref{fig:Ped_app_flow} demonstrates the flowchart of the app, from camera image collection to warning message display at the user interface. 

% \begin{figure}
% \centerline{\includegraphics[scale=.5]{figs/V_Ours/COSMOS_schema.jpg}}
% \caption{3D Schematic Diagram of the Proposed DT.}
% \label{fig:system_overall}
% \end{figure}

%\subsection{Use case: Intersection safety warning app prototype} 
% \sd{YJ/Mahshid:  
% Update the diagram, with ROI/end-user like in \cite{oliveira2024microservices} combining Fig. 1, 4, 5. building on your existing drawing, 
% also need to mention the toolboxes/algo. used in each step (AI tools: YOLO,CARLA...) to be consistent with the literature review}

\section{Evaluation}
\label{subsec:our_eval}

In safety-critical applications, we evaluate our approach from multiple perspectives. First, we measure the end-to-end system latency, which is critical for real-time safety warning deployment. Second, we assess the accuracy of warning message generation, validated through CARLA simulation experiments. In addition, we evaluate multi-user localization accuracy using UWB chips, and conduct user-end validation through VR experiments to assess system effectiveness and usability.

\subsection{End-to-end latency}

Measuring latency could be particularly challenging due to the need for cross-network timing synchronization of devices and compute server. Table~\ref{tab:pipeline-elem-latency} presents the latency incurred by the main components of a typical video analytics pipeline for object detection and tracking, as shown in Fig.~\ref{fig:safety-ROI-app}. We measured the latency for three sizes of YOLOv8 object detection model: YOLOv8s (small, $\sim$11.2 million parameters), YOLOv8m (medium, $\sim$25.9 million parameters), and YOLOv8x (large, $\sim$68.2 million parameters). Larger models offer improved detection accuracy but come at the cost of increased latency and higher GPU and memory usage~\cite{Jocher_YOLO_by_Ultralytics_2023}. The pipeline has been systematically optimized in terms of memory and resource usage. These elements usually run on an edge device or server. The results identify potential bottlenecks within the pipeline and indicate which components could benefit from further optimization. %Additionally, the table illustrates the impact of video content complexity on the latency of each component, providing insights into performance variations during \sd{busy and sparse traffic conditions (Mahshid, can we still tell sparse versus dense, given we removed them from Tabl VII...?}. \md{should I remove discussion about traffic condition or keep it in the text and discuss more?}\sd{yes, maybe just remove it?}

\begin{table*}[t]
\vspace{0.3cm}
\caption{Average latency and standard deviation for different pipeline elements.}
\centering
\resizebox{\textwidth}{!}{
\begin{tabular}{
    >{\raggedright\arraybackslash}m{2cm}|| % Metric column
    >{\centering\arraybackslash}m{1.5cm}| % Reception
    >{\centering\arraybackslash}m{2cm}| % Pre-processing
    >{\centering\arraybackslash}m{1.5cm}| % YOLOv8s
    >{\centering\arraybackslash}m{1.5cm}| % YOLOv8m
    >{\centering\arraybackslash}m{1.5cm}| % YOLOv8x
    >{\centering\arraybackslash}m{1.5cm}|| % Object Tracking
    >{\centering\arraybackslash}m{1.5cm}| % MQTT Msg Creator
    >{\centering\arraybackslash}m{1cm}| % MQTT Retrieval (Ethernet)
    >{\centering\arraybackslash}m{1cm}| % MQTT Retrieval (Wi-Fi)
    >{\centering\arraybackslash}m{1cm}| % MQTT Retrieval (LTE)
    >{\centering\arraybackslash}m{1cm}  % MQTT Retrieval (5G)
}
\hline
\multirow{3}{*}{\textbf{Metric}} & \multicolumn{6}{c||}{\textbf{Inference Elements}} & \multicolumn{5}{c}{\textbf{Downstream Elements}} \\ \cline{2-12}
 & \textbf{Reception} & \textbf{Pre-processing} & \multicolumn{3}{c|}{\textbf{Object Detection Model}} & \textbf{Object Tracking} & \textbf{MQTT Msg Creator} & \multicolumn{4}{c}{\textbf{MQTT Msg Retrieval}} \\ \cline{4-6} \cline{9-12}
 &  &  & \textbf{Small (YOLOv8s)} & \textbf{Medium (YOLOv8m)} & \textbf{Large (YOLOv8x)} &  &  & \textbf{Ethernet} & \textbf{Wi-Fi} & \textbf{LTE} & \textbf{5G} \\ \hline\hline
\textbf{Avg Latency (Std Dev) [ms]} & 1.94 (1.69) & 0.108 (0.024) & 4.034 (0.084) & 7.216 (0.086) & 11.140 (1.800) & 0.973 (0.173) & 0.081 (0.021) & 3.21 (0.315) & 6.86 (1.19) & 45.72 (15.30) & 39.21 (7.12) \\ \hline
\end{tabular}
}
\label{tab:pipeline-elem-latency}

\end{table*}

\subsection{Safety message generation accuracy}

\begin{figure}[h]
\centering
\includegraphics[width=1.0 \columnwidth]{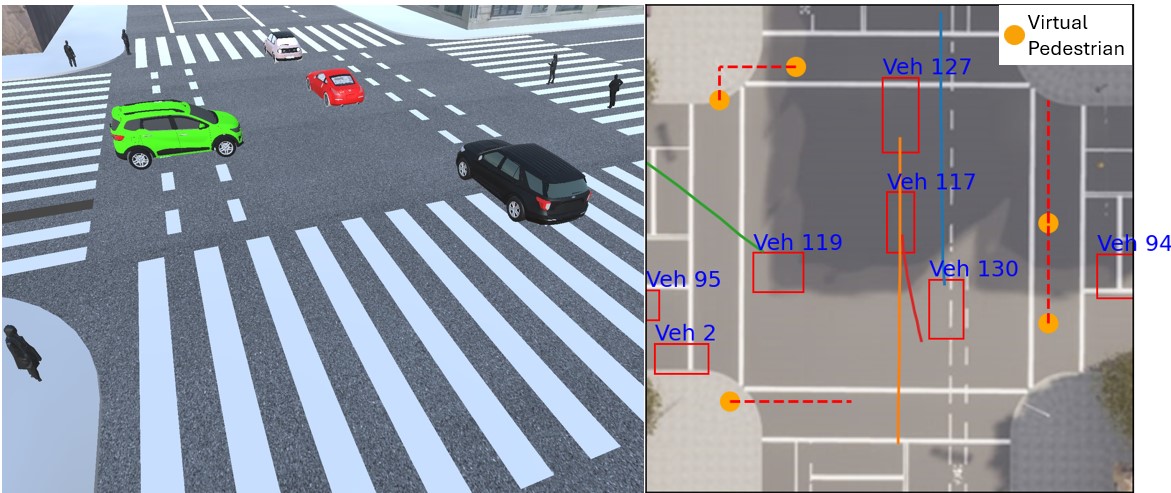}
\caption{A simulated pedestrian in CARLA (left) surrounded by cars with real-world positions (right).}
\label{fig:carla-validation}
\end{figure}

% We conduct an experiment in a simulation to validate the safety warning application. In the real world, near-crash cases are rare, so we generate $165$ virtual pedestrians in the simulation and test our application. We track the position of each pedestrian and regard it as a real collision if the distance between the pedestrian and any vehicle is less than 50 pixels in the detection image. 
To validate the accuracy of our trajectory prediction and risk assessment algorithms for warning generation, we conducted three rounds of simulation, each lasting $10$ minutes and generating a total of $232$ virtual pedestrians in CARLA. We then compared the number of collision warning messages with the actual number of simulated collisions. The process is illustrated in Fig.~\ref{fig:carla-validation}.

\begin{figure}[h]
\centering
% First subfigure (TTC first now)
\begin{subfigure}[b]{0.49\columnwidth}
\centering
\includegraphics[width=\textwidth]{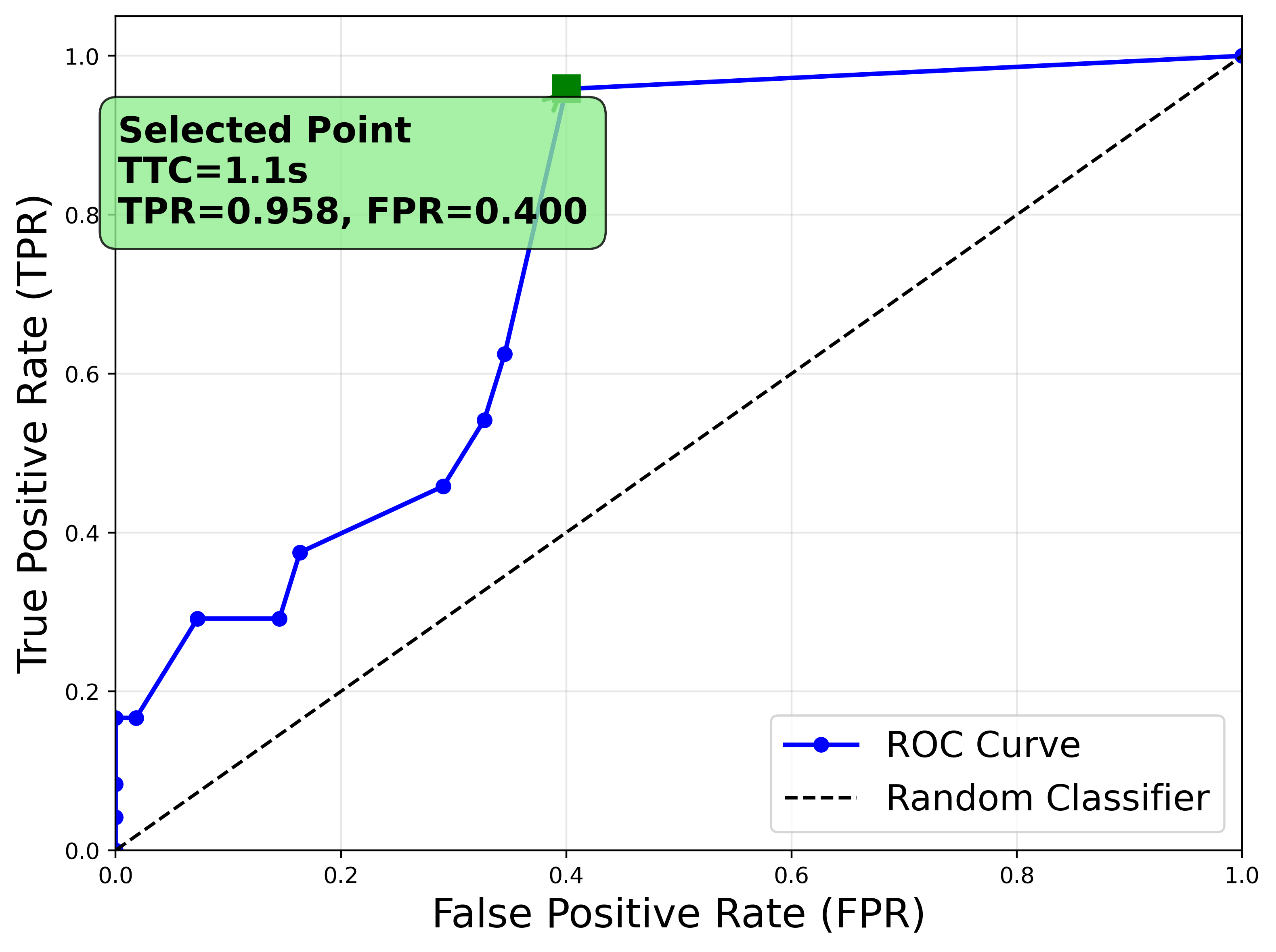}
\caption{TTC (in seconds)}
\label{fig:roc-ttc}
\end{subfigure}
% Second subfigure (Distance second now)
\begin{subfigure}[b]{0.49\columnwidth}
\centering
\includegraphics[width=\textwidth]{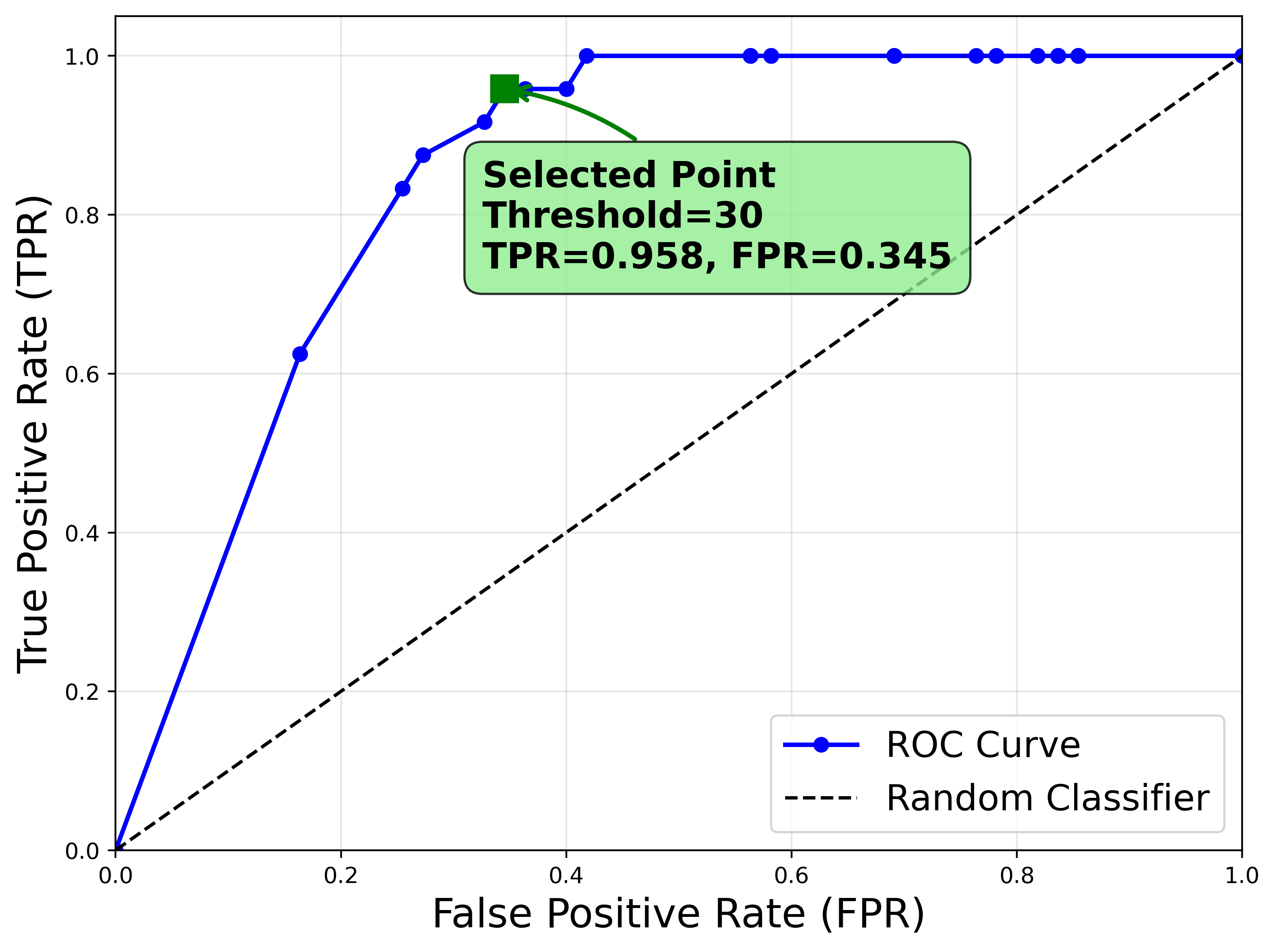}
\caption{Distance (in pixels)}
\label{fig:roc-danger-distance}
\end{subfigure}
\hfill % ensures horizontal spacing   
\caption{ROC curves for trajectory prediction to determine optimal thresholds}
\label{fig:roc-curve}
\vspace{-0.5cm}
\end{figure}

Note that metrics for risk assessment include 
TTC (i.e., the time remaining before a collision occurs) and post-encroachment time (PET) (i.e., the time interval between when the encroaching vehicle leaves the conflict point and when the vehicle of the right-of-way arrives at the conflict point). 
TTC is commonly used, while PET is typically for post-event analysis. %to assess safety levels. 
%Here we select the TTC with a threshold of $1.5s$ is usually selected for safety warning applications \cite{chen2013forward}. 
%The warning is generated by our risk assessment algorithm, which checks the next several predicted trajectories to see if they are close enough. Based on the predicted warning and the actual collision, 
To select the optimal thresholds of TTC, 
we first generate the ROC (Receiver Operating Characteristic) curve in Fig.~\ref{fig:roc-curve}. Fig.~\ref{fig:roc-ttc} illustrates the relationship between the true positive rate (TPR) and the false positive rate (FPR) across various TTC threshold values, ranging from 0.1 to 1.2 seconds. The optimal TTC threshold is 1.1 seconds (indicated by a blue square), at which the TPR reaches $0.958$ and the FPR is $0.4$. To compute TTC, we need to compare each pair of predicted trajectory points in the next $t$ time steps, determined by a danger distance threshold that is defined as when the proximity between a vehicle and a pedestrian constitutes a dangerous interaction.
% We define a distance threshold that determines when the predicted distance between the car and the pedestrian is considered dangerous for calculating TTC. 
Fig.~\ref{fig:roc-danger-distance} presents the TPR and FPR for the threshold of danger distances ranging from $5$ to $100$ pixels. 
The optimal threshold is valued at $30$ pixels, which yields a TPR of $0.958$ and an FPR of $0.345$. 
Based on the above two thresholds, we run the collision prediction model. 
%\sd{also help calibrate ped. traj. prediction and risk score $P((X-X^{v})_{t:t+3}/V<\alpha_{TTC})>\beta$?}
The resultant confusion matrix for predicted collisions under the selected thresholds is given by $[[\text{TP}, \text{FP}], [\text{FN}, \text{TN}]] = [[66, 45], [2, 119]]$.

% summarized in Table.~\ref{confusion-matrix}. 
%From the table, only $2$ false negative samples are generated, which represents not generating a warning for a possible collision with a VRU.

% \begin{table}[h]
% \begin{center}
% \begin{tabular}{|c|c|c|}
% \hline
%   & Predicted Positive & Predicted Negative \\
% \hline
% Actual Positive & $66$ & $2$ \\
% \hline
% Actual Negative & $45$ & $119$ \\
% \hline
% \end{tabular}
% \caption{Confusion Matrix for Predicted vs Actual Pedestrian Collision.}
% \label{confusion-matrix}
% \end{center}
% \end{table}
\subsection{UWB Localization Accuracy and Query Latency}
We evaluate the UWB-based multi-person localization module in terms of positioning accuracy and real-time query latency. Localization accuracy is measured using the mean Euclidean error between the estimated position $\hat{\mathbf{p}}_i$ and the ground-truth position $\mathbf{p}_i$:

\begin{equation}
e_i = \|\hat{\mathbf{p}}_i - \mathbf{p}_i\|_2 .
\end{equation}

Table~\ref{tab:uwb_accuracy} reports the mean localization error under single-user and two-user scenarios. In addition, we measure the measurement frequency, defined as the elapsed time between a user request and server-side position response. From the results, the system achieves accurate localization with an error within 10 cm. We implement the user-end system using a Python-based user interface with a CLI on each anchor end, which provides superior performance compared to the phone-based application. The low measurement frequency observed in the multi-user scenario is primarily due to the reconnection time required when switching between different users during successive queries. We utilize a constant-velocity model to interpolate positions between two measurements and estimate the localization error in the two-user scenario.

In real-world testing, we identified limitations of using UWB in outdoor environments. UWB signals can be blocked by moving objects, such as vehicles, and Wi-Fi signals may interfere with UWB performance \cite{stocker2025modeling}. This issue is especially noticeable at urban intersections with dense surrounding Wi-Fi signals, while rural intersections are less affected due to lower Wi-Fi interference.

% Add to preamble: \usepackage{multirow}
\begin{table}[h]
    \centering
    \caption{UWB Multi-User Localization Accuracy and Query Time}
    \begin{tabular}{p{3cm}p{2.0cm}p{2.2cm}}
    \toprule
    Scenario & Mean Error $\pm$ Std Dev  (cm) & Measure Frequency (Hz) \\
    \midrule
    % Single & \multirow{2}{*}{$7.983 \pm4.071$} & 100 \\
    Single & $7.983 \pm4.071$ & 10 \\

    Two Users  & $22.456\pm6.897$ & 0.3 \\
    \bottomrule
    \end{tabular}
    \label{tab:uwb_accuracy}
\vspace{-0.7cm}
\end{table}

% \begin{table}[h]
% \centering
% \caption{UWB Localization Query Latency \sd{has this fig. been used in PAVE? Make sure we also cite PAVE paper somewhere...}}
% \begin{tabular}{lcc}
% \toprule
% \# of Users & Mean Latency (ms) & Std Dev (ms) \\
% \midrule
% 1 User & 100 &  \\
% 2 Users & 1500 &  \\
% \bottomrule
% \end{tabular}
% \label{tab:uwb_latency}
% \end{table}

\subsection{Human response time assessment}

% \begin{figure}[h]
%     \centering
%     \includegraphics[width=0.33\textwidth]{figs/V_Ours/participant_demographics.jpg}
    
%     \caption{Participant demographics in VR experiments. \sd{IRB for extended age groups}}
% \label{fig:participant_demographics}
% \end{figure}

\begin{figure}[h]
    \centering
    \includegraphics[width=0.42\textwidth]{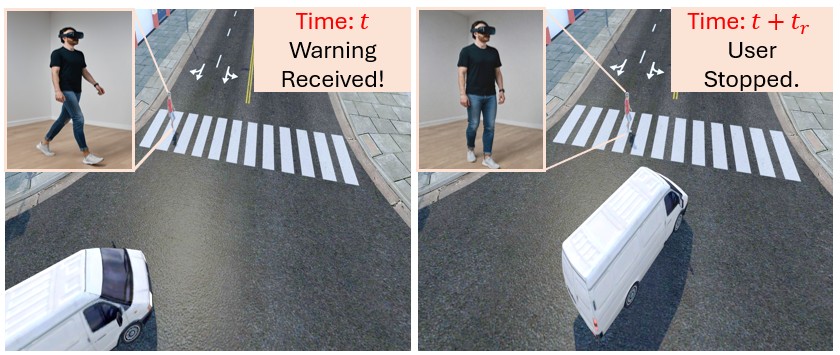}
    \caption{Response time determination in the VR experiment}
\label{fig:vr-exp}
\vspace{-0.5cm}
\end{figure}

\begin{figure}[h]
    \centering
    \includegraphics[width=0.37\textwidth]{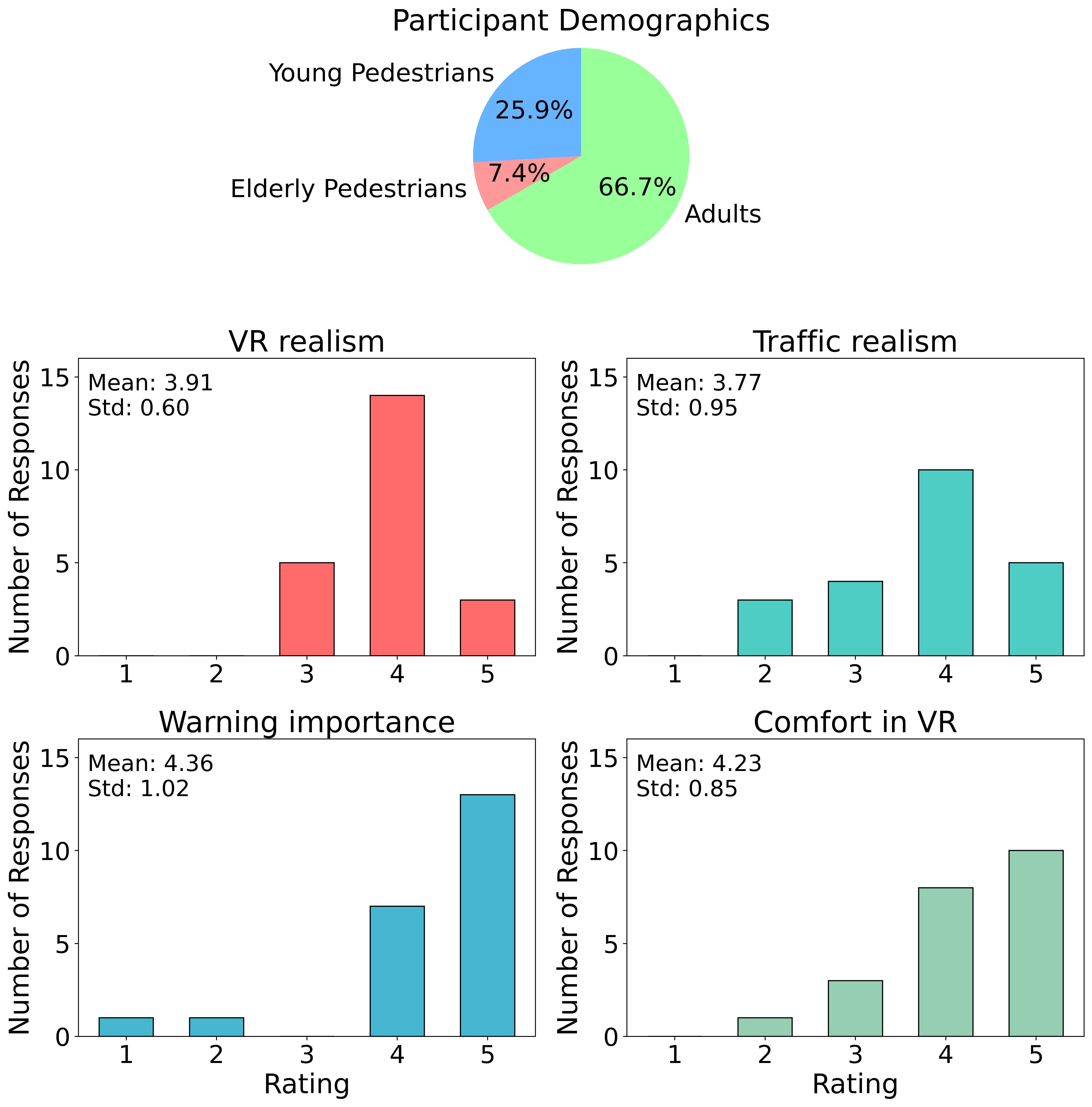}
    
    \caption{Survey results from the VR experiment categorize participants into three age groups: young ($<20$ years), adults ($20–65$ years), and elderly pedestrians ($>65$ years).}
\label{fig:participant_demographics}
\vspace{-0.5cm}
\end{figure}

% \begin{table}[ht]
% \centering
% \caption{Participant Demographics in the VR Experiment}
% \begin{tabular}{|l|c|}
% \hline
% \textbf{Participant Group} & \textbf{Number of Users} \\
% \hline
% Young Pedestrians & 7 \\
% Elderly Pedestrians & 2 \\
% Adults & 18 \\
% \hline
% \textbf{Total} & 26 \\
% \hline
% \end{tabular}
% \label{tab:participant_demographics}
% \end{table}

\begin{figure}%[h]
\centering
\includegraphics[width=0.9 \columnwidth]{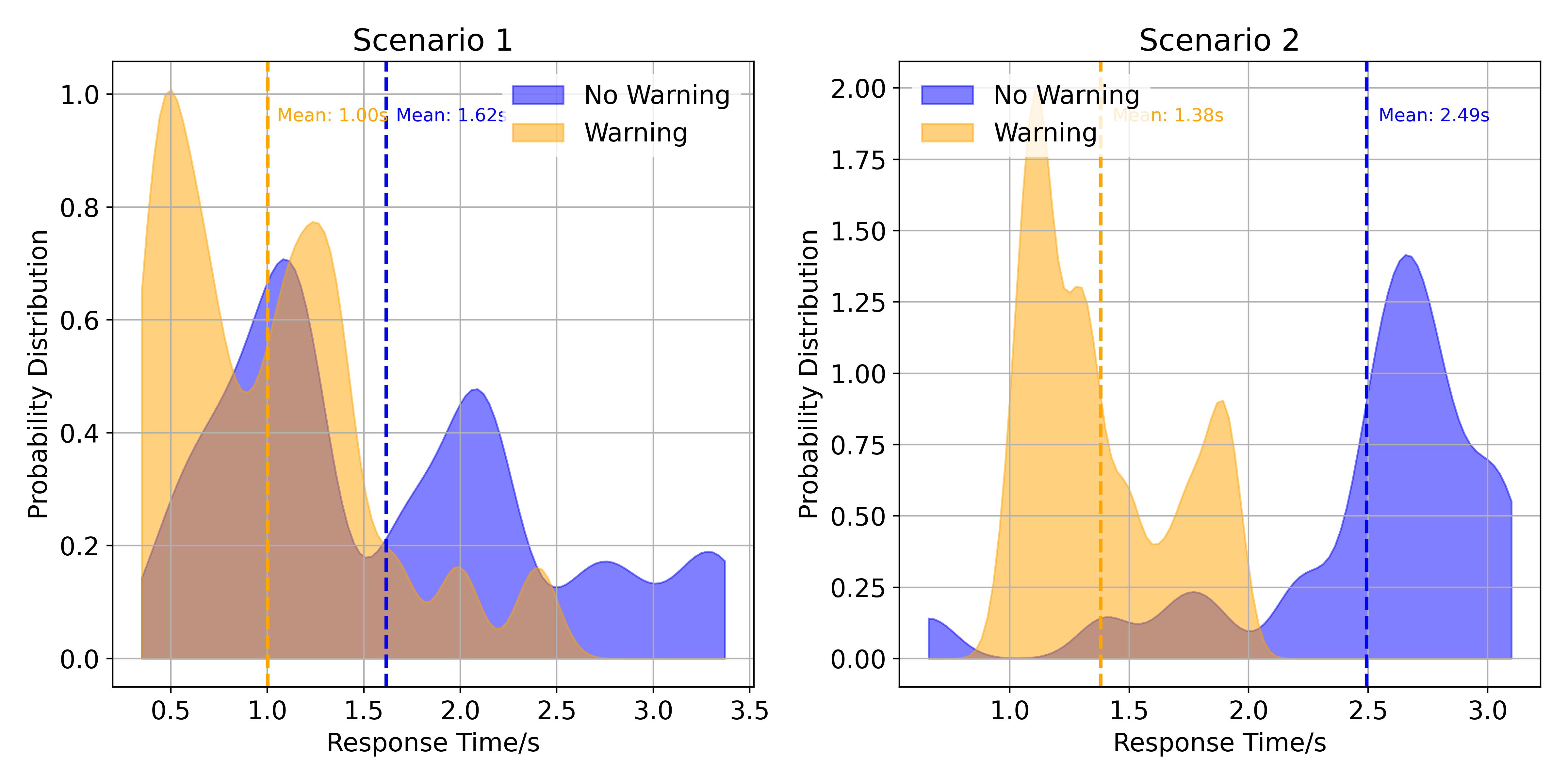}
\caption{
Comparison of response times with and without a warning. A scooter (in Scenario 1) and a vehicle (in Scenario 2) poses the danger, respectively.}
\label{fig:VR-exp}
\vspace{-0.3cm}
\end{figure}

Would issuing warnings to pedestrians help reduce users' response time and increase their safety awareness?
To test this hypothesis, we designed virtual reality (VR) experiments in Unity3D \cite{unity2024}, where warnings are provided via voice and text displayed on the VR headset Meta Quest 3. 
There are two traffic scenarios, 
one involving an interaction between the participant  (i.e., a pedestrian) and an oncoming scooter, 
and the other between the participant and an oncoming vehicle. 
In each scenario, participants first receive no warning and then a warning, and their response times are recorded. As illustrated in Fig.~\ref{fig:vr-exp}, when traffic approaches at time $t$, the user either receives a warning or does not. The user stops after a response time $t_r$, which is then recorded. %\sd{does it mean the same participant first receives a warning and the next does not? Or you randomize the participants who receive and who do not?} 
The response time distributions are presented in Fig.~\ref{fig:VR-exp}, with the `no warning' condition shown in yellow and the `warning' condition in blue. 
%In Scenario 1, the warning reduced the average response time by $0.62\mathrm{s}$, and in Scenario 2, by $1.11\mathrm{s}$.
The issuance of warning has reduced people's average response time by $0.62\mathrm{s}$ (in Scenario 1) and $1.11\mathrm{s}$ (in Scenario 2), respectively. The results demonstrate that warning messages sent to users help reduce their response time.
The demographics of the participants are shown in the upper section of Fig.~\ref{fig:participant_demographics}. 
The bar chart summarizes the participants’ survey responses regarding VR and traffic realism, the perceived importance of safety warnings, and comfort within the VR environment, with mean and standard deviation indicated on the top left corners of each subfigure. Based on the survey results, participants were generally satisfied with the realism of the VR environment and traffic conditions, and considered the warning messages important.
%Most responses were rated 4 or 5, indicating generally positive feedback across these aspects.

\section{Conclusions and Future Research}
\label{sec:conclud}

Our deployment on the COSMOS testbed in New York City demonstrates the generalizability of the proposed digital twin framework in dense, multimodal urban environments. Manhattan’s high traffic density, limited space, and diverse transportation modes provide a realistic and challenging setting for studying vehicle–VRU interactions. The COSMOS infrastructure enables large-scale data collection and live experimentation that would otherwise be difficult to achieve.

We design the system with scalability in mind. The sensing and analytics modules (e.g., camera, UWB) are resource-efficient, modular, and deployable across distributed edge and cloud servers. This architecture supports flexible workload distribution and facilitates extension from a single intersection to network-wide deployments. The framework is model-agnostic, allowing AI components to be replaced or adapted to different camera configurations and urban environments.

Rather than relying on customized infrastructure (e.g., dedicated C-V2X hardware), our pilot emphasizes leveraging legacy traffic cameras and existing wireless networks for cost-effective and rapid deployment. Future work will explore integration with emerging communication technologies such as C-V2X and advanced V2P protocols to further reduce latency and enhance reliability. Additional research directions include multi-intersection coordination, large-scale deployment evaluation, privacy-preserving communication, and robust multi-device synchronization in dense urban settings.

% % use section* for acknowledgment
% \section*{Acknowledgment}
% % This work is sponsored by NSF CPS-2038984. 
% We would like to thank Mengxuan Liu for her assistance in generating the 3D models. 

\ifCLASSOPTIONcaptionsoff
  \newpage
\fi



%\bibliographystyle{IEEEtran}

%\reftitle{References}

%\externalbibliography{yes}

\bibliographystyle{IEEEtran}

% \bibliography{IEEEabrv, ref/ref_Di,ref/ref_Fu,ref/ref_Ghasemi,ref/ref_Turkcan,ref/ref_Others,ref/ref_COSMOS}
\bibliography{ref/ref_Di,ref/ref_Fu,ref/ref_Ghasemi,ref/ref_Turkcan,ref/ref_Others,ref/ref_COSMOS,ref/ref_Abhi}

\end{document}